\documentclass[10pt,twocolumn,letterpaper]{article}

\usepackage{cvpr}
\usepackage{times}
\usepackage{epsfig}
\usepackage{graphicx}
\usepackage{amsmath}
\usepackage{amssymb}
\usepackage{subfigure} 
\usepackage{booktabs}
\usepackage{multirow}
\usepackage{lscape}
\usepackage{pdfpages}
\usepackage[breaklinks=true,bookmarks=false]{hyperref}

\cvprfinalcopy 

\def\cvprPaperID{10290} 

\ifcvprfinal\fi

\begin{document}

\title{MCEN: Bridging Cross-Modal Gap between Cooking Recipes and Dish Images with Latent Variable Model}

\author{Han Fu\textsuperscript{\rm $\dagger\ddagger$}\quad
	Rui Wu\textsuperscript{\rm $\dagger$}\quad
	Chenghao Liu\textsuperscript{\rm $\S$}\quad
	Jianling Sun\textsuperscript{\rm $\dagger\ddagger$\thanks{Corresponding author: Jianling Sun.}} \\
	\textsuperscript{\rm $\dagger$}Zhejiang University, Hangzhou, China\\
	\textsuperscript{\rm $\ddagger$}Alibaba-Zhejiang University Joint Institute of Frontier Technologies, China\\
	\textsuperscript{\rm $\S$}Singapore Management University, Singapore\\
	{\tt\small \{11821003, tactic, sunjl\}@zju.edu.cn,
	twinsken@gmail.com}\\
}

\maketitle
\thispagestyle{empty}

\begin{abstract}
   Nowadays, driven by the increasing concern on  diet and health, food computing has attracted enormous attention from both industry and research community. One of the most popular research topics in this domain is Food Retrieval, due to  its profound influence on health-oriented applications. In this paper, we focus on the task of cross-modal retrieval between food images and cooking recipes. We present Modality-Consistent Embedding Network (MCEN) that learns modality-invariant representations by projecting images and texts to the same embedding space. To capture the latent alignments between modalities, we incorporate stochastic latent variables to explicitly exploit the interactions between textual and visual features. Importantly, our method learns the cross-modal alignments during training but computes embeddings of different modalities independently at inference time for the sake of efficiency.
   Extensive experimental results clearly demonstrate that the proposed MCEN outperforms all existing approaches on the benchmark Recipe1M dataset and requires less computational cost.
\end{abstract}

\section{Introduction}

Food is the paramount necessity of human life. As the saying goes, \emph{we are what we eat}, food not only provides energy for life activities, but also plays a significant role in affecting human identity, social formation, history, and culture inheritance \cite{harris1998good}. In our daily life, food is intricately linked to people's convention, lifestyle, health and social activities. Nowadays, with the development of Internet and mobile applications, sharing recipes and food images on social platforms has become a widespread trend \cite{moed2007food}. Due to the massive amounts of data resource online, food computing has become a popular field, inciting numerous machine learning tasks such as ingredient recognition \cite{matsuda2012recognition,kagaya2014food}, food image retrieval \cite{wang2019learning} and recipe recommendation \cite{trattner2017investigating,sanjo2017recipe}. Among the research topics, Image-to-Recipe learning (im2recipe) is one of the most important problems due to its profound influence on health-oriented applications \cite{min2019survey}. For instance, food-health analysis applications are required to predict detailed nutrition contents and calorie information from food images, and a recipe-retrieval system is a necessary solution on this scenario. 

\begin{figure}
	\centering
	\subfigure[Prior Work]{\includegraphics[width=1.5in]{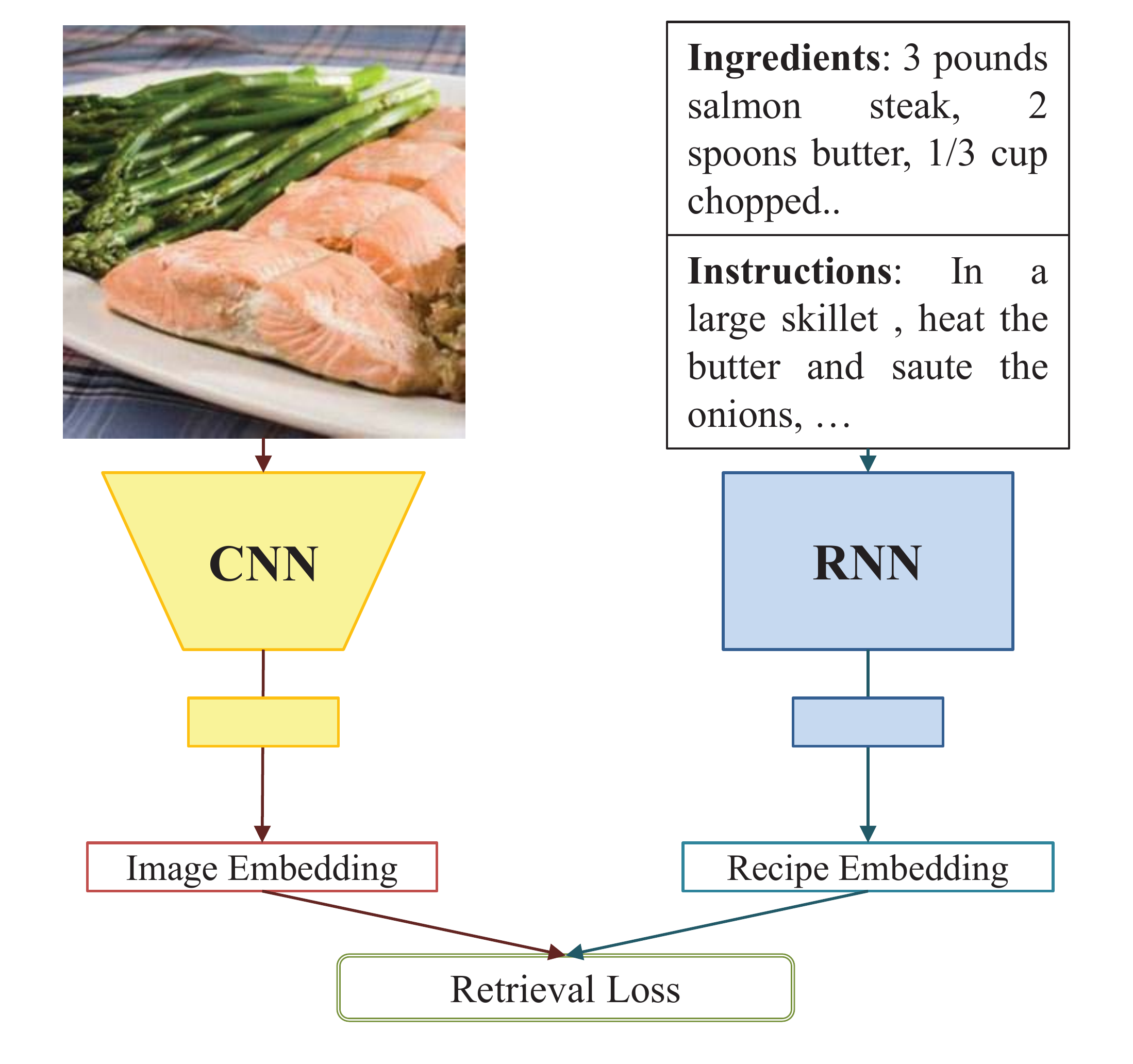}}
	\subfigure[Our Work]{\includegraphics[width=1.5in]{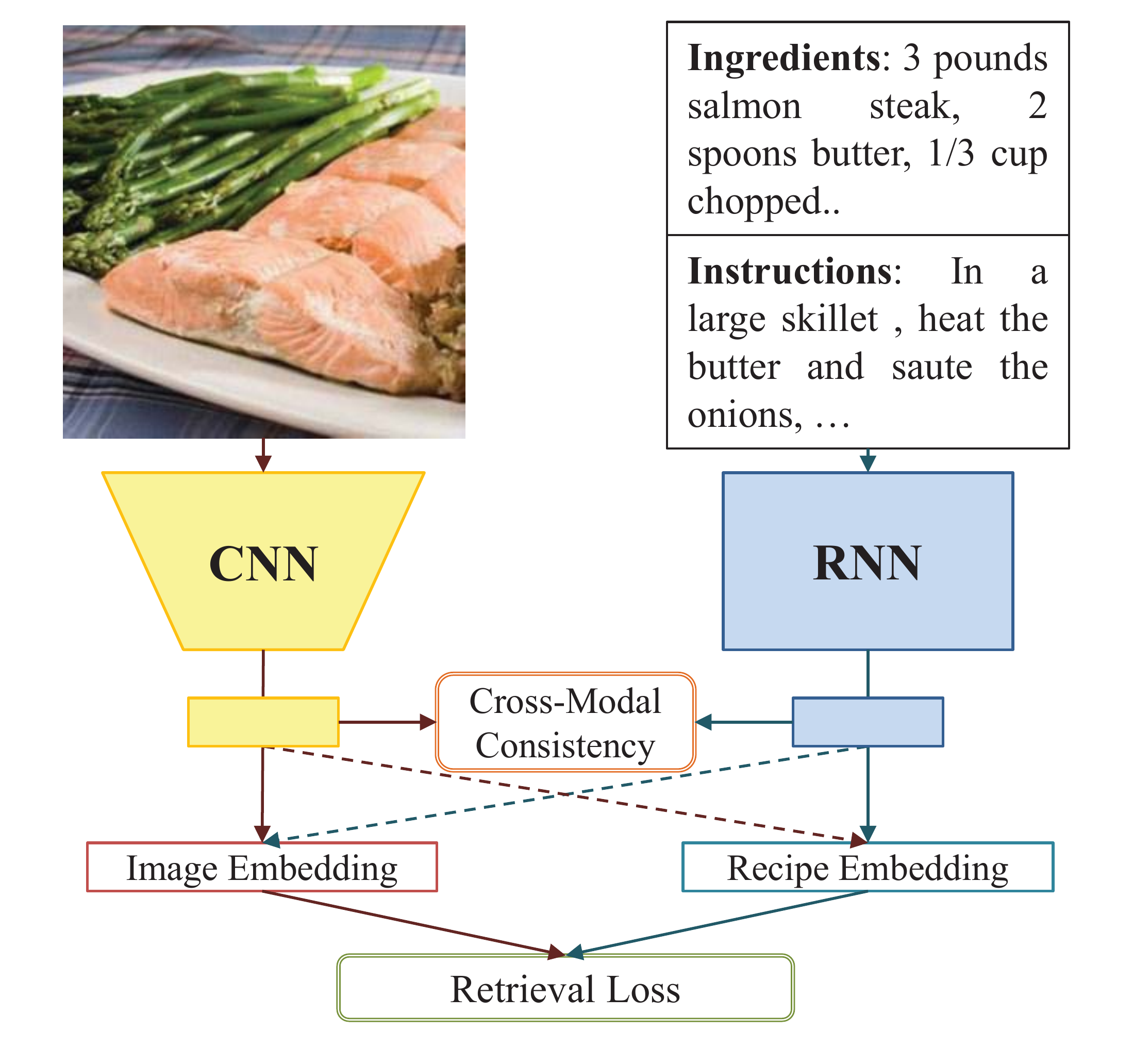}}
	\caption{\textbf{A comparison between prior work (a) and the proposed MCEN (b).} We learn modality-consistent embeddings by capturing the interactions between images and recipes via latent variables. The dotted lines represents that the joint information is only used during training. At inference time, the embeddings are computed independently.  }
	\label{intro}
\end{figure}

Im2recipe is a challenging task since it involves highly variant foods images and expatiatory textual recipes. A typical recipe consists of a list of ingredients and cooking instructions which may not directly align with the appearance of the corresponding food image. Typically, recent efforts have formulated im2recipe as a cross-modal retrieval problem \cite{salvador2017learning,marin2018recipe1m,carvalho2018cross,zhu2019r2gan}, to align matching recipe-image pairs in a shared latent space with retrieval learning approaches. Concretely, prior work builds two independent networks to encode textual recipes (ingredients and cooking instructions) and food images into embeddings respectively. And the retrieval loss object is learned to gather matching pairs and differentiate dissimilar items. Though existing methods are expressive and powerful, there remain two major concerns. 1) Current systems encode images and texts with two different networks independently. However, such independence brings barriers between modalities, resulting in obstacles to discover latent semantic alignments across modalities. Consequently, such approaches thus could suffer from polysemous instances \cite{song2019polysemous}. 2) The recipe representations are obtained based on fixed pre-trained skip-thought vectors \cite{kiros2015skip}, leading to highly diversities between textual and image feature spaces.

To alleviate such limitations, we strive to take a step towards capturing joint information of different modalities and injecting the cross-modal alignments into the embedding learning processes on both sides. We introduce \textbf{M}odality-\textbf{C}onsistent \textbf{E}mbedding \textbf{N}etwork (MCEN) which learns joint cross-modal representations for textual recipes and dish images. The major idea is to exploit the interactions between visual and textual features explicitly and share the cross-modal information to the embedding spaces of both modalities with stochastic latent variable models. The stochastic variable is leveraged to capture the latent correlations between modalities during training, while the embeddings can still be calculated independently at test time for high efficiency and flexibility. Moreover, The randomness introduced by latent variables is also beneficial for handling polysemous instances where one recipe corresponds with multiple images.

In a nutshell, the \textbf{main contribution} of this work is threefold:

\begin{itemize}
	\setlength{\itemsep}{2pt}
	\setlength{\parsep}{0pt}
	\setlength{\parskip}{0pt}
	\item We propose a novel cross-modal retrieval framework to obtain modality-consistent embeddings by explicitly capturing the correlations between recipes and food images with latent variables.
	\item We exploit the latent alignments during training with cross-modal attention mechanism and replace it with prior condition at inference time for efficiency.
	\item We propose a task-specific encoder for textual recipes based on hierarchical attentions, which cannot only adapt to the interaction with images, but also simplify and accelerate the training and inference procedure.
\end{itemize}

We conduct experiments on the challenging benchmark Recipe1M \cite{salvador2017learning} and the results demonstrate that our model significantly outperforms all state-of-the-art approaches on the cross-modal recipe retrieval problem and requires less computational overhead.

\section{Related Work}

\paragraph{Computational Cooking.}
Food and cooking are essential parts of human life, which are closely relevant to health \cite{trattner2017investigating}, social activities, bromatology, dietary therapy and culture \cite{harris1998good}, etc., profoundly affecting the quality of life. Therefore, research involving cooking recipes has drawn considerable attention. Food and cooking provide rich attributes on multiple channels, including both visual content (e.g., dish pictures) and texts (e.g., dish descriptions and cooking instructions). Current literature leverages the attributes in various ways. Typically, recent examples in computer visions are food classification and recognition \cite{chen2016deep,liu2016deepfood,lee2018cleannet,zhou2016fine,kagaya2014food}, and retrieval of captions \cite{engilberge2018finding,chen2018deep}, ingredients \cite{chen2017cross, chen2018deep} or recipe instructions \cite{carvalho2018cross, salvador2017learning,min2017being,min2017delicious} according to dish images, while researchers from natural language processing community usually focus on such applications as recipe recommendation \cite{trattner2017investigating,sanjo2017recipe}, aligning instructions with video and speech \cite{malmaud2015s}, recipe texts generation from flow graph \cite{mori2014flowgraph2text}, workflow generation from recipe texts \cite{yamakata2016method}, cooking action tracking \cite{bosselut2018simulating}, recipe representation \cite{malmaud2014cooking}, checklist recipe generation \cite{kiddon2016globally} and recipe-based question answering \cite{yagcioglu2018recipeqa, malmaud2014cooking}. Moreover, there is also some work using machine learning approaches to connect health with food attributes, such as prediction of nutrient \cite{kusmierczyk2016online} or energy \cite{meyers2015im2calories}, and healthy recipe recommendation \cite{elsweiler2017exploiting, trattner2017investigating, yang2017yum}. All these efforts contribute to the prosperity of food computation and understanding, bridging the gap between machine learning applications and people's daily life.

Recent introductions of large-scale food-related datasets have further accelerated the research improvements on food understanding. Considering the application purpose, the datasets can be categorized into two groups: food recognition \cite{bossard2014food,matsuda2012recognition} and cross-modal recipe retrieval \cite{salvador2017learning, marin2018recipe1m, min2017being, min2017delicious, chen2016deep, salvador2017learning}. We focus on recipe retrieval task in this paper, aiming at retrieval relevant cooking recipes with respect to the image query and vise versa. Typically, the datasets for retrieval generally incorporate both food images and other information such as ingredients, structured cooking instructions and flavor attributes. Among the datasets, Recipe1M \cite{salvador2017learning} is the most well curated large-scale dataset
with preprocessed English textual information and we evaluate the effectiveness of our method on it in this paper. 

\paragraph{Text-Image Retrieval.}
Our work is related to current approaches on multi-modal retrieval task, where the key problem is to measure the similarity between a text and an image. The major challenge of this issue lies in the modality-gap, which means that the feature spaces of different modalities largely diverse from each other. Text-image retrieval is at meeting point between computer vision and natural language communities, attracting research attentions over decades \cite{lew2006content}. Traditional approaches formulate this issue as either a language modeling task \cite{kiros2014multimodal} or a correlation maximization problems \cite{rasiwasia2010new,gong2014multi} using canonical correlation analysis (CCA) \cite{hotelling1936relations}. Recently, many efforts have been made to build end-to-end retrieval systems leveraging deep learning methods \cite{srivastava2012multimodal,andrew2013deep,feng2014cross,yan2015deep,peng2016cross}. Another avenue is to improve the triplet loss with hard negative mining \cite{schroff2015facenet}, such as \cite{frome2013devise,wang2016learning,faghri2017vse++}. 

Despite of the progress, the above approaches encode different modalities into independent feature spaces, suffering from modality gap between heterogenetic contents. To address this issue, recent works incorporate attention mechanism to capture the latent alignment relationships between words and different image regions \cite{huang2017instance,lee2018stacked,li2019visual, wang2019camp}. Though expressive, these methods require massive computational overhead during inference since the cross-modal attention scores between a query and each item in the reference set need to be calculated, limiting the scalability to large-scale retrieval scenario. In this paper, we leverage latent variables to incorporate cross-modal attention mechanism into retrieval tasks during training but maintain independent calculations for different modalities respectively at inference time.

Image-to-recipe is a newly proposed task and is formulated as a cross-modal learning task by recent efforts \cite{chen2017cross,salvador2017learning}, to retrieve the relevant recipes based on image queries. Following these settings, several inspiring methods have been introduced to improve the retrieval performance by using such techniques as additional textual feature \cite{chen2018deep}, semantic information \cite{carvalho2018cross} and adversarial learning \cite{zhu2019r2gan,wang2019learning}.

\section{Modality-Consistent Embedding Network}
\subsection{Overview}
In this section we introduce the methodology of the proposed Modality-Consistent Embedding Network (MCEN).

\textbf{Problem Formulation.} 
The aim of the proposed framework is to measure the similarity between food images and the relevant textual recipes. Formally, denote $\{\mathbf{v}^i,\mathbf{r}^i\}_{i=1}^N$ as a set of $N$ image-recipe pairs where an image $\mathbf{v}^i\in\mathbf{V}$ and a recipe $\mathbf{r}^i\in\mathbf{R}$. The notations $\mathbf{V}$ and $\mathbf{R}$ denote the visual and recipe spaces. It should be noted that one recipe corresponding to multiple images is allowed. A recipe $\mathbf{r}^{i}$ consists of a set of ingredients $\mathbf{X}^{ing, i}$ and a list of  cooking instructions $\mathbf{X}^{ins, i}$. An image $\mathbf{v}^i$ contains the appearance of a completed dish. Importantly, the ingredients and cooking instructions of a recipe may not directly align with the appearance of the matching image, which brings additional heterogeneity challenge compared to traditional cross-modal retrieval tasks.

Considering the information gap between modalities, we set our target to learn the mapping functions from observed data to the embedding distributions as $\mathbf{V}\rightarrow\mathbf{E}^{v}$ and $\mathbf{R}\rightarrow\mathbf{E}^{r}$, where $\mathbf{E}^{v}\in\mathbb{R}^{d}$ and $\mathbf{E}^{r}\in\mathbb{R}^{d}$ denote the distributions of $d$-dimensional image embedding and recipe embedding respectively, so that a picture is closer to the corresponding recipe than any other image in the latent space. 

\begin{figure*}
	\centering
	\includegraphics[width=17.5cm]{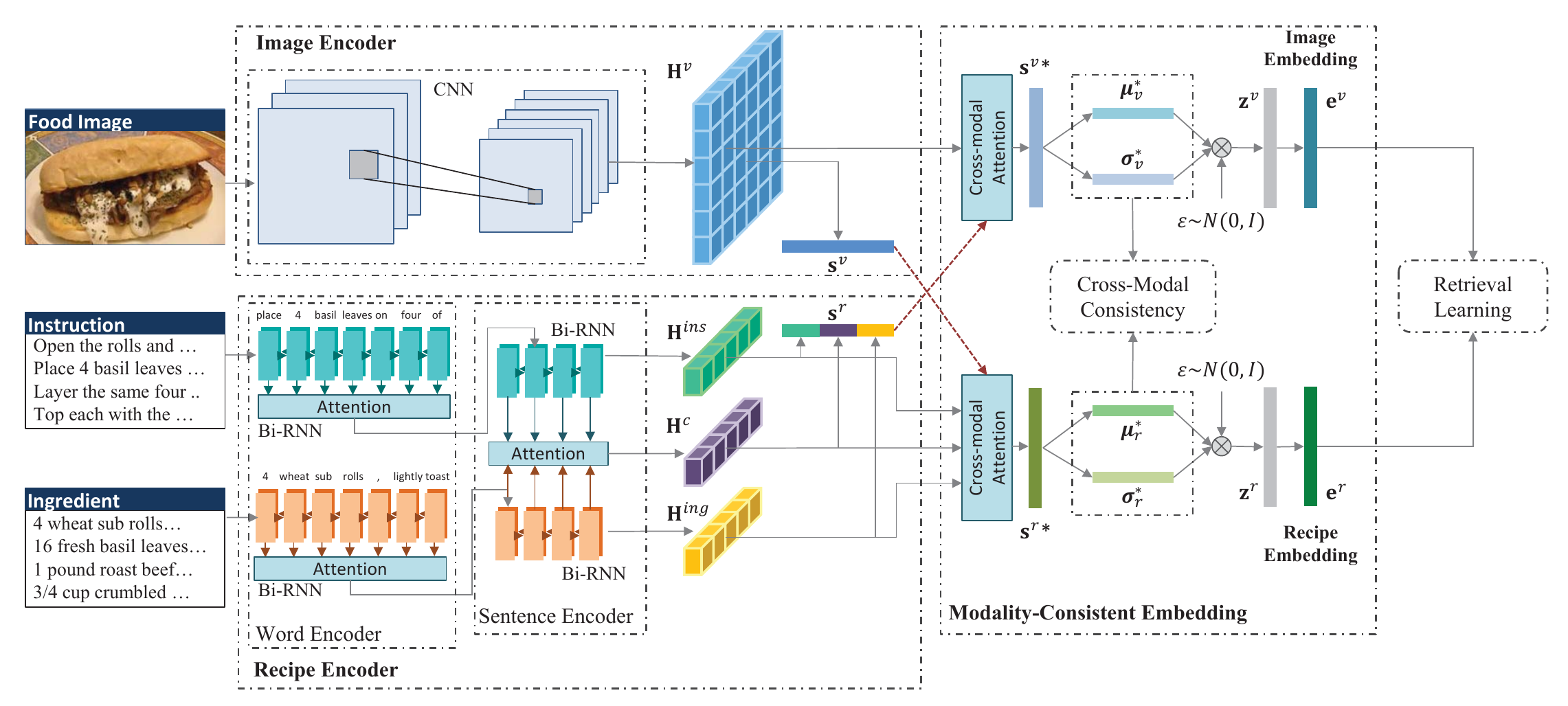}
	\caption{\textbf{The architecture and training flow of MCEN.} The red dotted lines denote that the cross-modal attention components only work during training and are omitted at testing time. The system is comprised of three major components: a recipe encoder, an image encoder, and a modality-consistent embedding component. The interaction between images and texts is captured with latent variables and shared by both latent spaces.}
	\label{arch}
\end{figure*}

\textbf{Architecture.}
The architecture of MCEN is illustrated in Figure \ref{arch}. The system consists of three major modules: a recipe encoder, an image encoder and an embedding learning component for modality-consistent space modeling. Through the training flow, the visual feature is extracted by feeding the food picture $\mathbf{v}^i$ to the CNN-based image encoder. Meanwhile, the high-level representations of instructions and ingredients are obtained by hierarchical attention-based RNN encoders. Then these representations are then fed to cross-modal attention components to exploit the interactions between images and texts. The cross-modal correlations are then leveraged to estimate the posterior distributions of embeddings with neural variational inference \cite{kingma2013auto,rezende2014stochastic}. With this method, we can discriminate training and inference process so as to reduce cross-modal computation at prediction time. To keep modality consistency, we align the distributions of latent representations by minimizing the KL-divergence of priors of different modalities. Finally, the latent representations sampled from the posterior distributions are passed to feed-forward layers to obtain the final embeddings of images and recipes respectively. The entire model is trained end-to-end with retrieval learning object.

The major novelty of MCEN comes from the incorporation of cross-modal correlation modeling with latent variables. MCEN captures the latent alignment relationships between images and texts during training while at inference time we do not require cross-modal attention since the posterior distribution is replaced by the prior during test. Though there exists prior work that focuses on modeling correlations between modalities \cite{lee2018stacked,li2019visual}, these approaches come with high computation overhead since the alignment score between a query and each reference instance needs computing as many times as the size of reference set \cite{song2019polysemous}. Conversely, MCEN obtains embeddings of different modalities independently during inference, which significantly reduces the computational overhead. Moreover, almost all prior methods require fixed pre-trained instruction vectors for recipes while parameters for image encoding are updated with respect to the retrieval object. The isomerism in training process leads to a diversity between feature spaces of images and recipes. In this work, the architecture of MCEN recipe encoder is quite different from prior systems and can be trained end-to-end from scratch.

\subsection{Image Encoder}
Given a food picture $\mathbf{v}$, the image encoder is responsible to extract the abstract features of the input. Different from previous methods, we use the output of the last residual block (res5c) of ResNet-50 \cite{he2016deep} which consists of $7 \times 7=49$ columns of 2048 dimensional convolutional outputs, denoted by $\mathbf{H}^{v}=(\mathbf{h}_1^{v},\mathbf{h}_2^{v},\cdots,\mathbf{h}_{49}^{v},)$. To obtain the representation for the image hidden states, we propose to use an attention layer, which estimates the importance of each hidden vector. Since a dish image may contain multiple objects that are not relevant to the recipe (i.e., forks and flowers), the aim of attention model is to force the encoder to focus more on regions that may contribute to the retrieval object. 

Formally, the image representation $\mathbf{s}^{v}$ is calculated with the weighted summation of convolutional states as:
\begin{equation}
	\mathbf{s}^{v}=\sum_{i=1}^{49}\alpha_{i}^{v}\mathbf{h}_{i}^{v},
\end{equation}
where $\alpha_{i}^{v}$ is the attention score at position $i$, representing the importance of this region, calculated by:
\begin{equation}
	\alpha_{i}^{v}={\rm softmax}(\mathbf{v}_v^\top{\rm tanh}(\mathbf{W}_v\mathbf{q}_v + \mathbf{U}_v\mathbf{h}_{i}^{v})),
\end{equation}
where $\mathbf{W}_v$, $\mathbf{U}_v$ and $\mathbf{v}_v$ are trainable matrices and vector. $\mathbf{q}_v$ is the attention query vector. Here, it is a trainable vector initialized from scratch. For sake of writing convenience, we call such attention layer as \emph{Attention Pooling} and the input annotations ($\mathbf{H}^{v}$) as \emph{Attention Context}.

\subsection{Recipe Encoder}
In the recipe branch, ingredients and instructions are encoded separately with similar networks. Since the ingredients or instructions of a recipe usually comprise multiple sentences, we use a hierarchical attention-based model to extract textual features. Each instruction/ingredient is first fed to a word-level bi-directional recurrent neural network (bi-RNN) with gated recurrent unit (GRU) \cite{cho2014learning} and the final word-level representations are calculated with attention pooling mechanism (Equation 1-2) where the RNN hidden states are used as the attention contexts. Denote $\mathbf{H}^{ins}=(\mathbf{h}^{ins}_{1}\cdots\mathbf{h}^{ins}_{m})$ and $\mathbf{H}^{ing}=(\mathbf{h}^{ing}_{1}\cdots\mathbf{h}^{ing}_{n})$ as the feature sequences of instructions and ingredients respectively, where $m$ and $n$ are the numbers of instructions and ingredients of a recipe, and each element $\mathbf{h}^{ins}_{t}/\mathbf{h}^{ing}$ is the abstract representation of an instruction/ingredient. To model the correlations between instructions and ingredients, we employ the attention-based RNN decoder \cite{bahdanau2014neural}, which takes $\mathbf{H}^{ins}$ as the sequential input and $\mathbf{H}^{ing}$ as the contexts respectively. The output of the RNN decoder is denoted as $\mathbf{H}^c=(\mathbf{h}_1^c,\cdots,\mathbf{h}_{m}^c)$ which contains the joint information of both instructions and ingredients. Then, $\mathbf{H}^c$, $\mathbf{H}^{ins}$ and $\mathbf{H}^{ing}$ are fed to independent sentence-level bi-RNNs and attention pooling layers to obtain the sentence-level representations, denoted as $\mathbf{s}^c$, $\mathbf{s}^{ins}$, and $\mathbf{s}^{ing}$ respectively. The final feature representation of the recipe is obtained by concatenating the three sentence representations as:
\begin{equation}
	\mathbf{s}^r=[{\mathbf{s}^{c}}^\top, {\mathbf{s}^{ins}}^\top, {\mathbf{s}^{ing}}^\top]^\top.
\end{equation}

\subsection{Modality-Consistent Embedding}
It is challenging to align feature representations of multiple modalities when the features are extracted with independent networks. To alleviate this issue, we incorporate latent variables to capture the interactions between modalities. This method converts the embedding computation into a generative process. Taking the image side for instance, the probability to generate a specific embedding $\mathbf{e}^v$ for a given image $\mathbf{v}$ is modeled as:
\begin{equation}
	p(\mathbf{e}^v|\mathbf{v})=p(\mathbf{e}^v|\mathbf{z}^v,\mathbf{v})p(\mathbf{z}^v|\mathbf{v}),
\end{equation}
where the latent vector $\mathbf{z}^v$ is assumed to capture the correlations between $\mathbf{v}$ and the corresponding recipe $\mathbf{r}$. The posterior of $\mathbf{z}^v$ should hence be conditioned on both the recipe $\mathbf{r}$ and image $\mathbf{v}$, denoted as $p(\mathbf{z}^v|\mathbf{v},\mathbf{r})$. The prior of latent variables is usually formulated as a standard Gaussian distribution, which may reduce the effectiveness in generation \cite{dai2019diagnosing}. Here we propose to estimate the prior distribution with a neural network model that jointly learns the prior knowledge and excavates cross-modal alignments based on single modality, denoted as $p(\mathbf{z}^v|\mathbf{v})$. To simplify the generative process, both prior and posterior distributions for latent variables are assumed to be Gaussian distributions. Concretely, the generative story is as follows. We sample a latent variable $\mathbf{z}^v$ from the prior Gaussian distribution as:
\begin{gather}
	\mathbf{z}^v|\mathbf{v}\sim\mathcal{N}(\boldsymbol{\mu}_v,{\rm diag}(\boldsymbol{\sigma}_v^2))\\
	\boldsymbol{\mu}_v=\mathbf{W}_\mu^v \mathbf{s}^v + \mathbf{b}_\mu^v\\
		\boldsymbol{\sigma}_v={\rm softplus}(\mathbf{W}_\sigma^v \mathbf{s}^v + \mathbf{b}_\sigma^v),
\end{gather}
where $\mathbf{W}_\mu^v$, $\mathbf{W}_\sigma^v$ and $\mathbf{b}_\mu^v$, $\mathbf{b}_\sigma^v$ are weight matrices and bias. Conditioned on the latent variable $\mathbf{z}^v$, we generate the final image embedding as:
\begin{equation}
	\mathbf{e}^v=f_v(\mathbf{z}^v),
\end{equation}
where $f_v$ is a mapping function implemented as a one-layer neural network with tanh activation.

Estimation of Equation 4 can be challenging since the distributions are intractable. We leverage neural variational inference \cite{kingma2013auto,rezende2014stochastic} to optimize the evidence lowerbound (ELBO) as:
\begin{equation}
	\mathbb{E}_{q(\mathbf{z}^v|\mathbf{v},\mathbf{r})}(\log p(\mathbf{e}^v|\mathbf{z}^v,\mathbf{v}))-D_{KL}(q(\mathbf{z}^v|\mathbf{v},\mathbf{r})\Vert p(\mathbf{z}^v|\mathbf{v})),
\end{equation}
where $D_{KL}(\cdot)$ is the Kullback-Leibler divergence and $q(\mathbf{z}^v|\mathbf{v},\mathbf{r})$ is the approximate posterior, estimated as:
\begin{gather}
\mathbf{z}^v|\mathbf{v},\mathbf{r}\sim\mathcal{N}(\boldsymbol{\mu}_v^*,{\rm diag}({\boldsymbol{\sigma}_v^*}^2))\\
\boldsymbol{\mu}_v^*=\mathbf{W}_\mu^{v*} \mathbf{s}^{v*} + \mathbf{b}_\mu^{v*}\\
\boldsymbol{\sigma}_v^*={\rm softplus}(\mathbf{W}_\sigma^{v*} \mathbf{s}^{v*} + \mathbf{b}_\sigma^{v*}),
\end{gather}
where $\mathbf{W}_\mu^{v*}$, $\mathbf{W}_\sigma^{v*}$ and $\mathbf{b}_\mu^{v*}$, $\mathbf{b}_\sigma^{v*}$ are trainable matrices and bias, which are independent from the prior model. The cross-modal representation $\mathbf{s}^{v*}$ is obtained with an attention pooling layer which takes the recipe representation $\mathbf{s}^r$ as the query vector and image region features $\mathbf{H}^v$ as the attention contexts. The lowerbound of the likelihood can be optimized by minimizing the triplet loss, formalized as:
\begin{equation}
	\mathcal{L}_{ret}^v=[s(\mathbf{e}^v_a, \mathbf{e}^i_n)-s(\mathbf{e}^v_a, \mathbf{e}^i_p)+m]_+
\end{equation}
where $s(\cdot)$ expresses the cosine similarity between two vectors, and $m$ is the margin of error. Subscripts $p$, $n$ and $a$ refer to positive, negative and anchor of a triplet respectively.

Cases are similar on the recipe side and the distinction lies in the calculation of cross-modal representation $\mathbf{s}^{r*}$ for posterior approximation $q(\mathbf{z}^r|\mathbf{v},\mathbf{r})$. Here, we obtain $\mathbf{s}^{r*}$ with the similar manner to $\mathbf{s}^r$ (Equation 3) but replace the original trainable query vector with the image feature $\mathbf{s}^i$.  Formally, the final retrieval learning object is defined as:
\begin{equation}
	\mathcal{L}_{ret} + \alpha\mathcal{L}_{KL},
\end{equation}
where $\mathcal{L}_{ret}$ is the summation of the triplet losses for image-to-recipe and recipe-to-image retrieval, and $\alpha$ is a trade-off hyper-parameter. $\mathcal{L}_{KL}$ is the summation of the KL divergences on both sides:
\begin{equation}
	\begin{split}
	\mathcal{L}_{KL} = D_{KL}(q(\mathbf{z}^v|\mathbf{v},\mathbf{r})\Vert p(\mathbf{z}^v|\mathbf{v})) + \\ D_{KL}(q(\mathbf{z}^r|\mathbf{v},\mathbf{r})\Vert p(\mathbf{z}^r|\mathbf{r})).
	\end{split}
\end{equation} 

Moreover, as discussed, we aim to align the distributions of both modalities. For this end, we simply push the prior embedding distributions of both modalities together by minimizing the following KL-divergence:
\begin{equation}
	\mathcal{L}_{cos}=D_{KL}(p(\mathbf{z}^v|\mathbf{v})\Vert p(\mathbf{z}^r|\mathbf{r})).
\end{equation}

\subsection{Cross-Modal Reconstruction}
Recent work \cite{zhu2019r2gan,wang2019learning} has proved the effectiveness of reconstruction loss on cross-modal recipe retrieval, since it encourages the embedding of one modality covers the corresponding information of the other modality. However, such an approach introduces additional network parameters to reconstruct the original images and recipes, which are too cumbersome for training a retrieval system. In this work we propose a much conciser method for cross-modal reconstruction. Instead of recovering the entire information of the original inputs, we only reconstruct the latent representations with the learned embeddings as:
\begin{gather}
	\mathbf{s}^{r'} = f^v_{r}(\mathbf{e}^v),\\
		\mathbf{s}^{v'} = f^v_{v}(\mathbf{e}^r),
\end{gather}
where $f^v_{r}$ and $f^v_{v}$ are mapping functions, implemented as two-layer neural networks. The formal reconstruction loss is formulated as:
\begin{equation}
	\mathcal{L}_{rec}=P(\mathbf{s}^{r'}, \mathbf{s}^{r}) + P(\mathbf{s}^{v'}, \mathbf{s}^{v}),
\end{equation}
where $P(\cdot)$ computes Pearson's correlation coefficient.

\subsection{Training and Inference}
The overall training object of MCEN is formulated as:
\begin{equation}
	\mathcal{L}=\mathcal{L}_{ret}+\alpha\mathcal{L}_{KL}+\beta\mathcal{L}_{cos}+\gamma\mathcal{L}_{rec},
\end{equation}
where $\alpha$, $\beta$ and $\gamma$ are hyper-parameters which balance the preference of different components.
The entire model can be trained end-to-end with the reparameterization trick \cite{kingma2013auto,rezende2014stochastic}. During inference, the latent variables are fixed to the expectation of prior distribution to stabilize the retrieval performance.

\section{Experiments}
\subsection{Settings}
\textbf{Dataset.}
The experiments are conducted on Recipe1M benchmark \cite{salvador2017learning}, a large-scale collection for recipe retrieval, including cooking instructions along with food images. The dataset consists of over 1M textual recipes and around 900K images. We use the same preprocessed samples provided by \cite{salvador2017learning} and we finally obtain 238,399 matching pairs of recipes and images for training, 51,119 pairs for validation and 51,303 pairs for test respectively. Moreover, it should be noted that we do not incorporate the additional semantic labels used by prior work \cite{salvador2017learning,carvalho2018cross,zhu2019r2gan}, such as food-classes and labels of commonly used ingredients.

\textbf{Metrics.}
We utilize the same metrics as the prior work \cite{salvador2017learning,carvalho2018cross,zhu2019r2gan}. Concretely, we compute median rank (MedR) and recall rate at top K (R@K) on sampled subsets in the test partition to evaluate the retrieval  performance. The sampling process is repeated for 10 times and the mean scores are reported. MedR measures the median retrieval rank position of true positives over all test samples, and the ranking position starts from 1. R@K refers to the percentage of queries for which matching instances are ranked among the top K results.

\begin{table*}
	\begin{center}
		\begin{tabular}{llcccccccc}
			\toprule
			\multirow{2}{0.6cm}{\textbf{Size}} & \multirow{2}{1.2cm}{\textbf{Methods}} & \multicolumn{4}{c}{\textbf{Image-to-Recipe}} & \multicolumn{4}{c}{\textbf{Recipe-to-Image}}\\
			\cline{3-10}
			& & \textbf{MedR} & \textbf{R@1} & \textbf{R@5} & \textbf{R@10} & \textbf{MedR} & \textbf{R@1} & \textbf{R@5} & \textbf{R@10} \\
			\midrule
			\multirow{8}{0.6cm}{1K} & Random & 500 & 0.1 & 0.5 & 1.0 & 500 & 0.1 & 0.5 & 1.0 \\
			& CCA \cite{salvador2017learning} & 15.7 & 14.0 & 32.0 & 43.0 & 24.8 & 9.0 & 24.0 & 35.0 \\
			& JE \cite{salvador2017learning} & 5.2 & 24.0 & 51.0 & 65.0 & 5.1 & 25.0 & 52.0 & 65.0 \\
			& ATTEN \cite{chen2018deep} & 4.6 & 25.6 & 53.7 & 66.9 & 4.6 & 25.7 & 53.9 & 67.1 \\
			& AdaMine \cite{carvalho2018cross} & 2.0 & 39.8 & 69.0 & 77.4 & 2.0 & 40.2 & 68.1 & 78.7 \\
			& R${\rm^{2}}$GAN \cite{zhu2019r2gan}& 2.0 & 39.1 & 71.0 & 81.7 & 2.0 & 40.6 & 72.6 & 83.3 \\
			& ACME \cite{wang2019learning} & 2.0 & 44.3 & 72.9 & 81.7 & 2.0 & 45.4 & 73.4 & 82.0 \\
			& MCEN (ours) & \textbf{2.0}$\pm 0.0$ & \textbf{48.2}$\pm 0.9$  & \textbf{75.8}$\pm 1.1$ & \textbf{83.6}$\pm 0.9$ & \textbf{1.9}$\pm 0.3$ & \textbf{48.4}$\pm 1.0$ & \textbf{76.1}$\pm 0.9$ & \textbf{83.7}$\pm 1.1$ \\
			\midrule
			\multirow{6}{0.6cm}{10K} & JE \cite{salvador2017learning} & 41.9 & - & - & - & 39.2 & - & - & - \\
			& ATTEN \cite{chen2018deep} & 39.8 & 7.2 & 19.2 & 27.6 & 38.1 & 7.0 & 19.4 & 27.8 \\
			& AdaMine \cite{carvalho2018cross} & 13.2 & 14.9 & 35.3 & 45.2 & 12.2 & 14.8 & 34.6 & 46.1 \\
			& R${\rm^{2}}$GAN \cite{zhu2019r2gan}& 13.9 & 13.5 & 33.5 & 44.9 & 12.6 & 14.2 & 35.0 & 46.8 \\
			& ACME \cite{wang2019learning} & 10.0 & 18.1 & 39.9 & 50.8 & 9.2 & 20.1 & 41.5 & 51.9 \\ 
			& MCEN (ours) & \textbf{7.2}$\pm0.4$ & \textbf{20.3}$\pm 0.3$ & \textbf{43.3}$\pm 0.3$ & \textbf{54.4 }$\pm 0.2$ & \textbf{6.6}$\pm 0.5$ & \textbf{21.4}$\pm 0.3$ & \textbf{44.3}$\pm 0.3$ & \textbf{55.2}$\pm 0.3$ \\
			\bottomrule
		\end{tabular}
	\end{center}
	\caption{\textbf{Retrieval Results of baselines.} The cross-modal retrieval performance is evaluated with MedR (lower is better) and R@K (higher is better). It should be noted that we do not incorporate pretraining embeddings and additional food-class labels which are utilized by prior approaches.}
	\label{main_result}
	\vspace{-1mm}
\end{table*}

\textbf{Implementation.}
For the image encoder, ResNet-50 \cite{he2016deep} pretrained on ImageNet \cite{deng2009imagenet} is used as the initialization weight. On the recipes side, the dimension of all hidden states is set to 300. Different from prior work, we do not use pretrained word embeddings. The entire recipe encoder is trained from scratch and the trainable parameters are initialized uniformly between $[-0.02, 0.02]$.

The dimension of final embeddings and all hidden states for neural inference is 1024. The margin of error $m$ is 0.3 and the hyper-parameters $\alpha$, $\beta$, $\gamma$ are set to 0.1, 0.002 and  0.008 respectively. The norm of gradient is clipped to be between $[-5, 5]$. We employ Adam solver \cite{kingma2014adam} with $\beta_1=0.9$, $\beta_2=0.999$ and $\epsilon=10^{-8}$ as the optimizer and the corresponding initial learning rate is set to $10^{-4}$. The model is trained end-to-end with batch-size 32. 

To train the model efficiently, we utilize two training strategies. First, as it is observed by other work \cite{bowman2016generating}, the loss for sequence modeling suffers from KL-divergence vanishing. To address this issue, we initialize $\alpha$ as $10^{-4}$ and gradually increase it to 0.1 as the training progress runs. Moreover, incorporating two independent stochastic variables can reduce the convergence speed. We therefore leverage a stage-wise strategy. Specifically, we fix the latent representation on the image side $\mathbf{z}^{r}$ as the mean of prior $\boldsymbol{\mu}^{r}$ and focus on training the recipe part. Then we alternatively train the posterior parameters on the image side after several epochs. Finally, early stopping strategy is applied and the model with best R@1 score on validation set is selected for testing.

\textbf{Comparison.}
The proposed MCEN is compared against several SOTA approaches:
\begin{itemize}
	\setlength{\itemsep}{1pt}
	\setlength{\parsep}{1pt}
	\setlength{\parskip}{1pt}
	\item CCA \cite{hotelling1936relations}, the Canonical Correlation Analysis method. The results are from \cite{salvador2017learning}.
	\item JE \cite{salvador2017learning}, a method to learn the joint embedding space of images and texts with pairwise cosine loss. This method also incorporates the classification task as a regularization.
	\item ATTEN \cite{chen2018deep}, a hierarchical attention model for cross-modal recipe retrieval. This approach also incorporates title information to extract recipe features.
	\item AdaMine \cite{carvalho2018cross}, a two-level retrieval approach which injects the semantic information into the triplet object.
	\item R${\rm^{2}}$GAN \cite{zhu2019r2gan}, a GAN-based method which learns cross-modal retrieval and multi-modal generation simultaneously.
	\item ACME \cite{wang2019learning}, the state-of-the-art method on cross-modal recipe retrieval task, which improves modality alignment using multiple GAN components. In our experiments, we use the released pre-trained model and report the results on our sampled test set.
\end{itemize}

\subsection{Main Results}
The main results on cross-modal retrieval task are listed in Table \ref{main_result}. Generally, the proposed MCEN consistently outperforms all baselines with obvious margin across all evaluation metrics and test sets. On the 1K set, MCEN achieves 2.0 median rank, which matches the SOTA results. In terms of R@K, MCEN achieves promising performance, beating all baselines including the to-date best approach ACME across all metrics on both image-to-recipe and recipe-to-image tasks.

On the 10K setting, the performances of all models decrease significantly since the retrieval task becomes much harder. As the size of subset increases, the gap between MCEN and previous methods becomes larger. Compared with the SOTA ACME method, our model achieves almost 30\% improvements on MedR metric over both im2recipe and recipe2im tasks, indicating the robustness of MCEN. 
 
\subsection{Ablation Studies}
\begin{figure}
	\begin{center}
		\includegraphics[width=8cm]{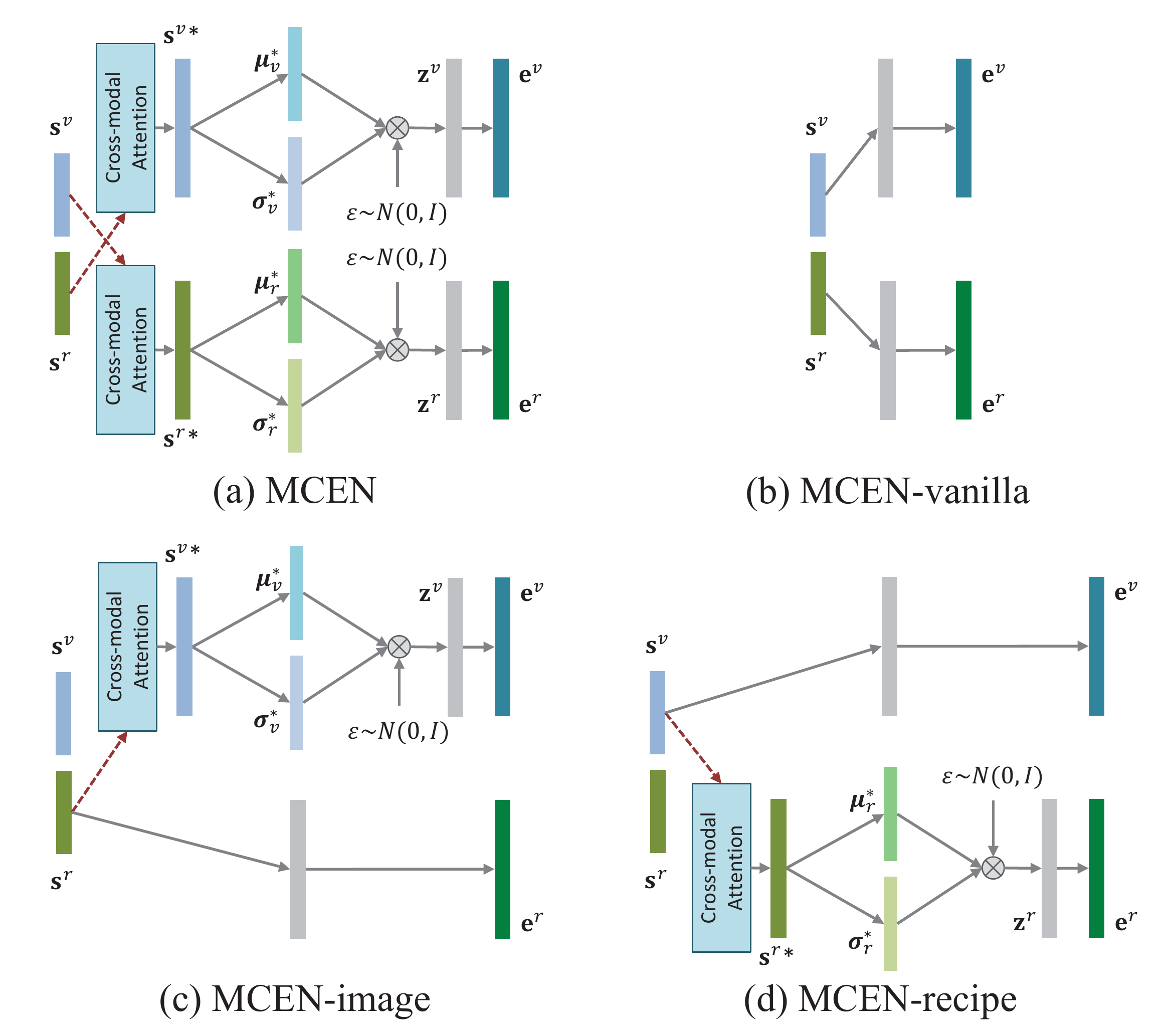}
	\end{center}
	\caption{Variants of architectures derived from MCEN.}
	\label{ablation}
	\vspace{-2.1mm}
\end{figure}

To evaluate the contributions of different components, we conduct ablation study on several variants of architectures detailedly. We depict  the variants of MCEN in Figure \ref{ablation}. MCEN-vanilla (Figure \ref{ablation} (b)) is the simplest architecture which does not incorporate any latent variables. The final embeddings $\mathbf{e}^r$ and $\mathbf{e}^v$ are obtained by:
\begin{align}
	\mathbf{e}^{r} &= g_r(\mathbf{s}^{r}),\\
	\mathbf{e}^{v} &= g_v(\mathbf{s}^{v}),
\end{align}
where $\mathbf{s}^{r}$ and $\mathbf{s}^{r}$ are the output of recipe encoder (Equation 3) and image encoder (Equation 1) respectively. The mappings $g_r$ and $g_v$ are implemented as two-layer neural networks with tanh activations. We also propose two variant models which leverage latent variables on either image (Figure \ref{ablation} (c)) or recipe side (Figure \ref{ablation} (d)). Besides, the performance of MCEN without reconstruction component (Equation 17-18) is also reported. For all variants derived from MCEN, the modality-consistency loss (Equation 15) is removed.

The retrieval results of different variant models on 1K subset are listed in Table \ref{tab_ab}. Not surprisingly, MCEN outperforms all variants with all evaluation metrics. It can be observed that the performance of MCEN-vanilla is similar to ACME (Table 1), indicating the effectiveness of the proposed architecture of the recipe encoder. Moreover, an interesting finding is that MCEN-image outperforms MCEN-recipe. A possible reason could be that, compared with rigmarole instructions, the relative semantic weights of different regions in an image are easier to be exploited.

\begin{table*}
	\begin{center}
		\begin{tabular}{lcccccc}
			\toprule \multirow{2}{2cm}{\textbf{Methods}} & \multicolumn{3}{c}{\textbf{Image-to-Recipe}} & \multicolumn{3}{c}{\textbf{Recipe-to-Image}}\\
			\cline{2-7}
			& \textbf{R@1} & \textbf{R@5} & \textbf{R@10} & \textbf{R@1} & \textbf{R@5} & \textbf{R@10} \\
			\midrule
			MCEN-vanilla & 44.5 & 72.3 & 80.7 & 44.9 & 72.8 & 80.9 \\
			MCEN-recipe  & 45.8 & 73.1 & 81.3  & 46.1 & 73.3 & 81.5 \\
			MCEN-image & 47.6 & 75.1 & 83.0 & 47.8 & 75.4 & 83.3\\
			MCEN w/o reconstruction & 46.4 & 75.4 & 83.1 & 47.8 & 75.7 & 83.3 \\
			MCEN & \textbf{48.2}  & \textbf{75.8} & \textbf{83.6} & \textbf{48.4} & \textbf{76.1} & \textbf{83.7}\\
			\bottomrule
		\end{tabular}
	\end{center}
	\caption{\textbf{Ablation Study}. The models are evaluated in terms of R@K with 1K subset.}
	\label{tab_ab}
\end{table*}

\begin{table}
	\begin{center}
		\begin{tabular}{lccc}
			\toprule
			\multirow{2}{1.5cm}{\textbf{Methods}} & \multirow{2}{1.2cm}{\textbf{\#Para}} & \multicolumn{2}{c}{\textbf{Speed}} \\
			\cline{3-4}
			& & \textbf{Train} & \textbf{Test}\\
			\midrule
			AdaMine \cite{carvalho2018cross} &  46.3M & 117.8  & 197.9 \\
			R${\rm^{2}}$GAN \cite{zhu2019r2gan} & 89.9M & 30.3 & 195.4 \\
			ACME \cite{wang2019learning} & 98.6M & 30.7 & 111.7 \\
			\midrule
			MCEN-vanilla & 48.9M & 57.6 & 194.9\\
			MCEN-recipe & 59.3M & 45.0 & 189.1 \\
			MCEN-image & 59.3M & 45.2& 188.7 \\
			MCEN & 69.6M & 42.7 & 185.8 \\
			\bottomrule
		\end{tabular}
	\end{center}
	\caption{\textbf{Statistics of parameters, training and testing speed (pairs/second).} All models are evaluated with the same settings on a single Titan XP GPU with batch-size 32. This comparison could be unfair since all the baselines require additional computational overhead for pre-training skip-thought vectors.}
	\label{speed}
	\vspace{-2mm}
\end{table}

\subsection{Analysis}
\paragraph{Parameters and Speed.}
We list the numbers of parameters and speeds of different systems in Table \ref{speed}. We can observe that although the inference network on either side introduces about 10.4M parameters, the additional parameters do not significantly decrease the training and test speed. Compared with the current SOTA ACME \cite{wang2019learning}, MCEN contains about 30\% less parameters and generates cross-modal embeddings with almost double speed, proving the high efficiency of the proposed architecture. The major reason for the gap between MCEN and ACME is that ACME requires additional overhead for adversarial learning.

\begin{figure}
\begin{center}
	\includegraphics[width=8.5cm]{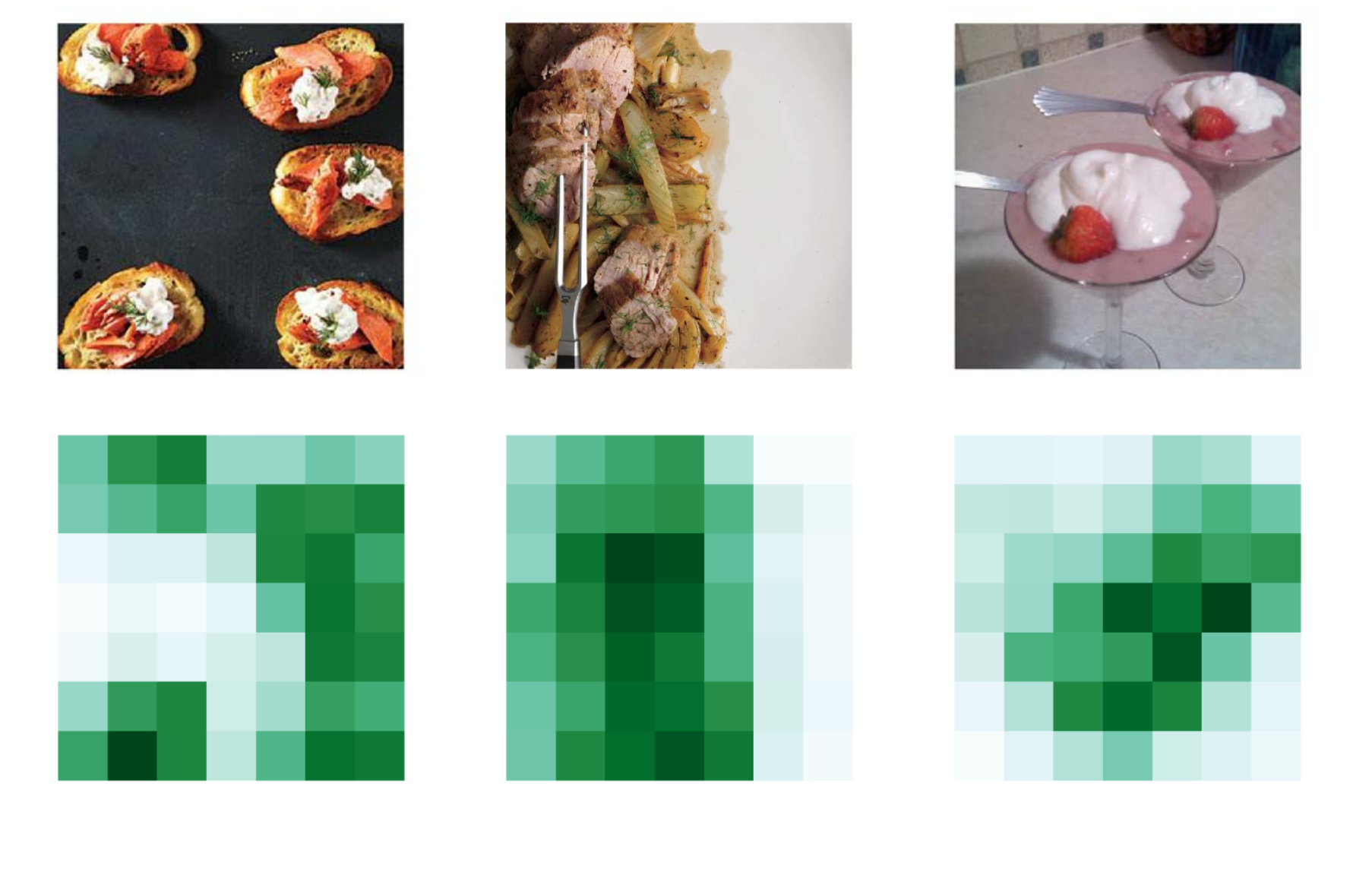}
	\caption{\textbf{Attention map of sampled images.} The darker color, the higher attention score.}
	\label{fig_att}
	\vspace{-3mm}
\end{center}
\end{figure}

\paragraph{Effectiveness of Cross-modal Attention.}
To better understand what has been learned by the cross-modal attention components, we visualize the intermediate results with attention. As shown in Figure \ref{fig_att}, the attention model learns to focus more on the valid regions containing food and ignore the background. Consequently, the final image embeddings are more constrained and not likely to be affected by noises (i.e. fork and  tablecloth) or polysemous instances.

\begin{figure}
	\begin{center}
		\includegraphics[width=8.5cm]{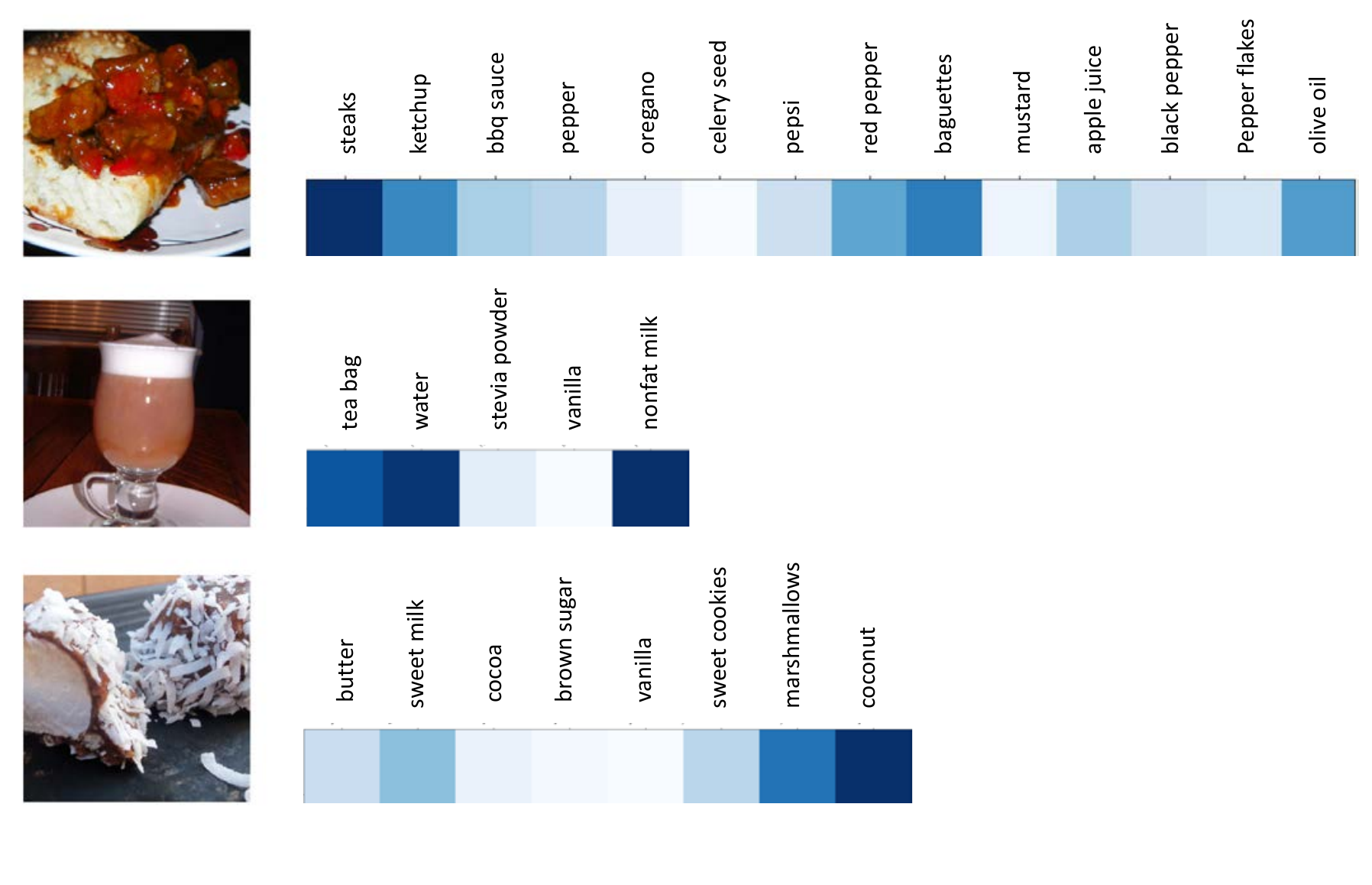}
		\caption{\textbf{Visualization of ingredient attention.} The model focuses on important ingredients with high attention scores.}
		\label{ingr_att}
		\vspace{-3mm}
	\end{center}
\end{figure}

On the recipe side, as shown in Figure \ref{ingr_att}, the attention model learns to focus on ingredients which can be interpreted based on visual connections with the food images. Taking the first sub-picture in Figure \ref{ingr_att} for instance, the attention model attaches highest weights to the three ingredients: \emph{steak}, \emph{ketchup} and \emph{baguettes}, which make up nearly the entire dish.
These observations demonstrate that the proposed MCEN learns to capture the semantic alignment relationships between images and recipes.

\section{Conclusion and Future Work}
In this paper, we propose a Modality-Consistent Embedding Network, namely MCEN, for cross-modal recipe retrieval. The proposed model focuses on modeling the interactions between food images and textual recipes during training with latent variables. Concretely, the latent variables are modeled based on cross-modal attention mechanisms during training while the embeddings of different modalities are still calculated independently during inference. We conduct experiments on the challenging Recipe1M dataset and the evaluation results with different metrics demonstrate the efficiency and effectiveness of MCEN. In the future, we are interested in incorporating pretrained language models into cross-modals analysis tasks.

\section*{Acknowledge}
We would like to thank the reviewers for their detailed comments and constructive suggestions.

{\small
\bibliographystyle{ieee_fullname}
\bibliography{cvpr}
}

\clearpage


\section*{Supplementary material}

# Supplementary Material for "MCEN: Bridging Cross-Modal Gap between Cooking Recipes and Dish Images with Latent Variable Model"

## A. Training Details

The image encoder is initialized with the pretrained ResNet-50 of PyTorch implementation [1]. During training, the images are resized to 256 pixels in their shortest side and random crops of 224 × 224 are taken. During inference, we simply use the central 224 × 224 pixels. We use dropout on top of the encoder output (Equation 1) and the final output layer (Equation 8) with dropout rate 0.2.

On the recipe side, we keep a maximum of 20 ingredients and 25 instruction sentences per recipe. Moreover, we truncate each sentence to a maximum of 30 words. The implementation of attention-based RNN decoder follows the classic implementation Dl4mt [2], where the previous hidden state is fed to a GRU before being fed to the decoder. Dropout is also applied with the similar strategy as the image encoder.

During training, we first freeze the parameters of ResNet and only optimize the rest parameters for 30 epochs. Then we unfreeze the ResNet weights and fine-tune the entire model for another 100 epochs. Early stopping strategy is utilized to select the best model with R@1 on validation for both periods.

## B. Scalability and Stability

We compare the MedR score of MCEN and other baselines against subsets larger than 10K to investigate the scalability of our model. As shown in Figure 1, it can be observed that MCEN outperforms all baselines on all test sets. Moreover, as the size of test set increases, the performance gap between MCEN and baselines becomes larger, indicating the robustness of MCEN. Especially on the largest 50K set, MCEN achieves a significant improvement, ranking the true positive by 15.3 positions ahead compared with ACME.

Furthermore, we also list the results of MCEN and some baselines on different sampled subsets. We can observe that the R@1 scores of MCEN on different test sets are more stable than those of ACME and Adamine. This comparison proves that the proposed model is more robust than the

[1] https://download.pytorch.org/models/resnet50-19c8e357.pth
[2] (https://github.com/nyu-dl/dl4mt-tutorial)

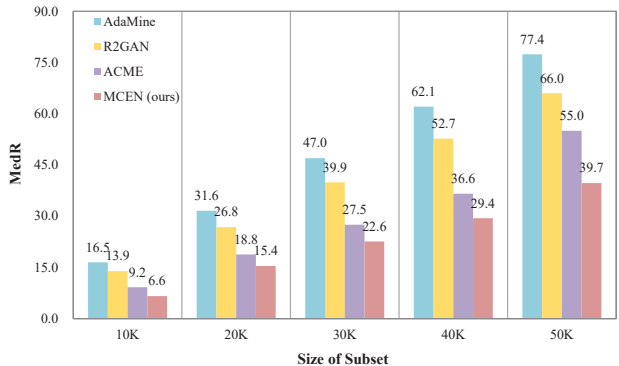

Figure 1. Scalability comparison between different models for image-to-recipe retrieval. Median Rank (MedR, lower is better) is used as the evaluation metric.

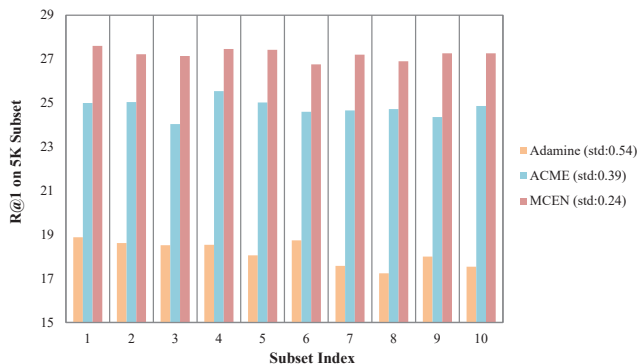

Figure 2. Stability comparison between ACME and MCEN (ours) on im2recipe task. Recall at 1 (R@1, higher is better) is used as the evaluation metric.

baseline and is capable to adapt to various settings.

## C. Analysis of Learned Embedding

We depict the learned image embeddings obtained by MCEN using a *t*-SNE visualization in Figure 3. The items are sampled from 5 of the most frequent classes and the color of each point indicates the category it belongs to. It can be observed that the embeddings with the same label are quite close while items in different classes are relatively

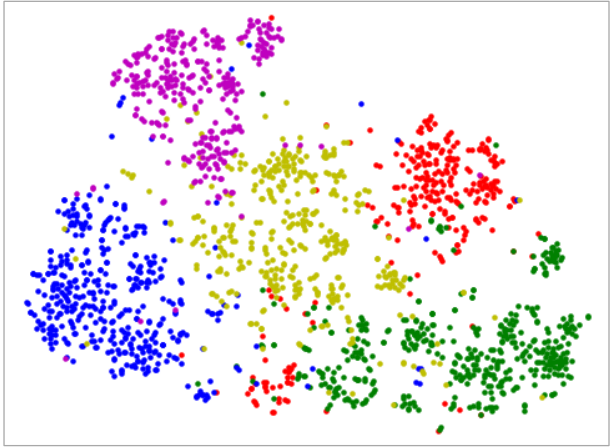

Figure 3. *t*-SNE visualization of learned embeddings.

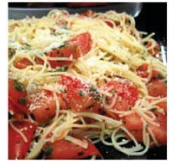

**True Ingredients**: pasta, tomatoes, olive oil, parmesan cheese, parsley, garlic, basil, salt
**Retrieved Ingredients**: chicken breasts, olive oil, lemon juice, pepper, salt, parmesan cheese, pasta, onion, tomatoes, basil, parsley, mozarella cheese

**True Insturctions**: Cook pasta according to package directions. In a bowl , combine remaining ingredients. Rinse and drain pasta and add to tomato …
**Retrieved Insturctions**: Combine all seasoning for chicken except mayonaise and lemon juice on a plate. Brush chicken lightly with mayonnaise / lemon …

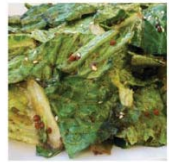

**True Ingredients**: lettuce, romaine, garlic, ginger, pepper, sesame oil, soy sauce, water, sesame seeds
**Retrieved Ingredients**: lettuces, oakleaf, romaine, olive oil, garlic, salt, pepper, lemon, water, lemon juicey

**True Insturctions**: Cut romaine crosswise into 2-inch pieces and put in a bowl. Cook garlic, ginger, and red-pepper flakes in sesame oil in a small skillet …
**Retrieved Insturctions**: Put 4 salad plates in the freezer to chill. Assemble all the ingredients as well as a garlic press, a big salad bowl, and a strainer …

Figure 5. Sampled results of image-to-recipe retrieval on 50K test set.

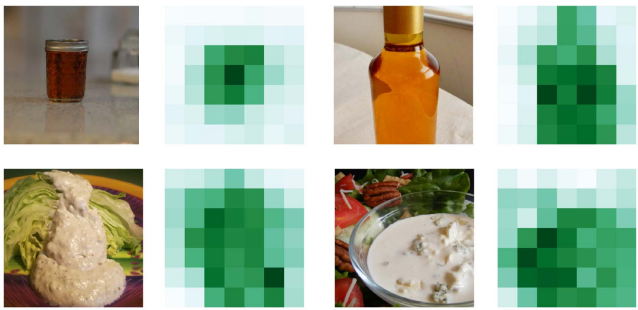

Figure 4. Attention visualization of polysemous cases, where multiple images (on the same line) correspond to one recipe.

far away from each other.

An interesting observation of Recipe1M dataset is the high variations in images where multiple images may correspond with single recipe since dishes cooked by diverse users can be very different. However, current methods do not explicitly address this problem. As shown in Figure 4, the proposed cross-modal attention approach is capable to force the model to focus more on common and important regions shared by various images. With this technique, images corresponding to the same recipe can be mapped to similar embeddings.

## D. Case Study

We visualize some retrieval results of MCEN on im2recipe and recipe2im tasks in Figure 5 and Figure 6 respectively. For image-to-recipe, we can find that the retrieved recipe has several common ingredients with the true one. For recipe-to-image task, it can be observed that the retrieved images are visually similar to the ground truth one.

**Ingredients:** 2 medium acorn squash, 4 tbsp. olive oil, kosher salt, pepper, 1 c apple cider, 1 tbsp red wine vinegar, 1 tbsp . whole - grain mustard…
**Insturctions:** Boil the potatoes in salted water for about 12 minutes, drain and keep aside. Heat the ghee in a wok over a medium flame. When hot add the mustard …

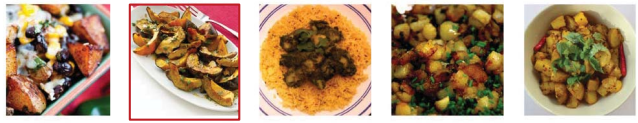

**Ingredients:** 1 Lb turkey sausage, the breakfast kind 12 large eggs, 12 teaspoon salt, 12 teaspoon black pepper, 1 cup cheddar cheese , shredded, 1 cup salsa …
**Insturctions:** In a large non-stick pan, crumble and cook turkey until no longer pink, about 8 minutes. Remove from pan. Wipe pan clean with paper towel and …

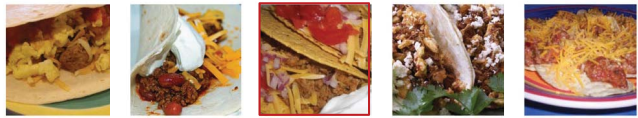

Figure 6. Sampled results of recipe-to-image retrieval on 50K test set. The top-5 retrieved images are shown from left to right, and the ground truth image is boxed in red.



\end{document}